\documentclass{article}

\usepackage{PRIMEarxiv}

\usepackage[utf8]{inputenc} 
\usepackage[T1]{fontenc}    
\usepackage{hyperref}       
\usepackage{url}            
\usepackage{booktabs}       
\usepackage{amsfonts}       
\usepackage{nicefrac}       
\usepackage{microtype}      
\usepackage{lipsum}
\usepackage{fancyhdr}       
\usepackage{graphicx}       
\graphicspath{{media/}}     
\usepackage{tabularray}
\usepackage{caption}
\title{Machine Learning in Proton Exchange Membrane Water Electrolysis – Part I: A Knowledge-Integrated Framework

}

\author{
  Xia Chen \\
  Sustainable Building Systems \\
  Leibniz University Hannover \\
  Hannover, Germany\\
  \texttt{xia.chen@iek.uni-hannover.de} \\
  \AND
  Alexander Rex \\
  Institute of Electric Power Systems \\
  Leibniz University Hannover \\
  Hannover, Germany\\
  \texttt{rex@ifes.uni-hannover.de} \\
  \And
  Janis Woelke \\
  Institute of Electric Power Systems \\
  Leibniz University Hannover \\
  Hannover, Germany\\
  \texttt{woelke@ifes.uni-hannover.de} \\
  \And
  Christoph Eckert \\
  Institute of Electric Power Systems \\
  Leibniz University Hannover \\
  Hannover, Germany\\
  \texttt{eckert@ifes.uni-hannover.de} \\
    \And
  Boris Bensmann \\
  Institute of Electric Power Systems \\
  Leibniz University Hannover \\
  Hannover, Germany\\
  \texttt{boris.bensmann@ifes.uni-hannover.de} \\
    \And
  Richard Hanke-Rauschenbach \\
  Institute of Electric Power Systems \\
  Leibniz University Hannover \\
  Hannover, Germany\\
  \texttt{rhr@ifes.uni-hannover.de} \\
    \And
  Philipp Geyer \\
  Sustainable Building Systems \\
  Leibniz University Hannover \\
  Hannover, Germany\\
  \texttt{philipp.geyer@iek.uni-hannover.de} \\
}

\begin{document}
\maketitle

\begin{abstract}
In this study, we propose to adopt a novel framework, Knowledge-integrated Machine Learning, for advancing Proton Exchange Membrane Water Electrolysis (PEMWE) development. Given the significance of PEMWE in green hydrogen production and the inherent challenges in optimizing its performance, our framework aims to meld data-driven models with domain-specific insights systematically to address the domain challenges. We first identify the uncertainties originating from data acquisition conditions, data-driven model mechanisms, and domain expertise, highlighting their complementary characteristics in carrying information from different perspectives. Building upon this foundation, we showcase how to adeptly decompose knowledge and extract unique information to contribute to the data augmentation, modeling process, and knowledge discovery. We demonstrate a hierarchical three-level framework, termed the "Ladder of Knowledge-integrated Machine Learning," in the PEMWE context, applying it to three case studies within a context of cell degradation analysis to affirm its efficacy in interpolation, extrapolation, and information representation. This research lays the groundwork for more knowledge-informed enhancements in ML applications in engineering.
\end{abstract}


\section{Introduction}
\label{sec:Introduction}

The integration of Machine Learning (ML) with domain-specific knowledge is a pivotal advancement in predictive modeling \cite{karpatne2017physics, raissi2019physics}. This combination has brought a new level of precision and insight to fields within engineering and environmental sciences \cite{ding2023machine, willard2022integrating}. While the synergy has notably improved accuracy and decision-making processes \cite{hippalgaonkar2023knowledge, von2021informed}, the challenge of seamlessly blending domain knowledge with ML algorithms continues to evolve. To bridge this gap, the Ladder of Knowledge-integrated Machine Learning has been introduced \cite{chen2023pathway}. This framework aims to optimize the utilization of domain-specific insights, offering a comprehensive approach to integrating prior knowledge information into ML applications.

Inspired by the long debate between holistic and reductionist approaches in ML \cite{minsky1991logical}, the framework aims firstly to synergize multidisciplinary domain knowledge with data-driven processes in two principal dimensions: firstly, by identifying and understanding the complementary nature of uncertainties in data, knowledge-based methodologies, and data-driven methods; secondly, by exploring knowledge decomposition from various perspectives and aligning these insights with our paradigm. Finally, building upon the previous two foundations in the specific domain context, the ladder unfolds across three progressive levels of integrating domain expertise into ML approaches \cite{chen2023pathway}.

In the pursuit of sustainable energy solutions, Proton Exchange Membrane Water Electrolyzers (PEMWEs) stand out for their high energy efficiency and minimal environmental impact \cite{grigoriev2020current} in hydrogen production. However, PEMWEs face significant challenges, including high capital costs\cite{mayyas2019manufacturing, saba2018investment}, component degradation\cite{feng2017review, chen2023key}, and complex modeling demands \cite{ma2021comprehensive, kumar2019hydrogen}. The lack of comprehensive datasets further complicates these challenges for developing interpretable ML applications in the domain, highlighting the need to integrate domain knowledge into ML for PEMWE advancements. While ML models have achieved breakthroughs in material properties and operational parameters discovery \cite{briceno2021machine}, deeper exploitation of electrochemical reactions and material behavior understanding \cite{chen2022advances, ma2021comprehensive} 
 is yet to be fully realized in ML models. Other domain-specific insights, such as efficiency factors \cite{makhsoos2023perspective}, scalability \cite{immerz2018effect}, and life cycle assessment \cite{gulotta2022life}, each could uniquely integrate and contribute to the development of advanced ML methods. We've observed similar research conducted in various energy domains, e.g., the lithium-ion accumulators \cite{thelen2022integrating, tu2023integrating}; Yet, a systematic integration framework to reconcile and utilize domain-specific knowledge with data-driven methods in PEMWEs development is missing. 

In a nutshell, the Ladder of Knowledge-integrated Machine Learning enables engineers to efficiently pinpoint their domain expertise and integrate them into a data-driven approach from various perspectives \cite{chen2023pathway}: Starting at the fundamental level, this level focuses on using prior knowledge in data augmentation and feature engineering to achieve better model performance within the given data range. The second level expands the models’ predictive capabilities for extrapolation and generalization. The third level sheds light on the logic behind predictions for knowledge discovery, enhanced information representation, and explainability toward informed decision-making. While integrating expert knowledge within ML has precedent in diverse fields, the Ladder of Knowledge framework distinguishes itself by its systematic approach and adaptable scalability.

Combining with case studies, our work aims to harmonize the information utilization gap between extensive domain expertise and cutting-edge ML methods, thus propelling the field into knowledge-integrated innovation in improving the ML practical application. To the best of our knowledge, this study is the first to systematically discuss the integration of domain-specific knowledge with ML in the context of PEMWEs. The novelty of this research lies in providing such an overview that paves the way for groundbreaking improvements in PEMWE research and applications in the broader field of energy systems. 

\begin{figure}[h]
	\centering
	\includegraphics[width=12.5cm]{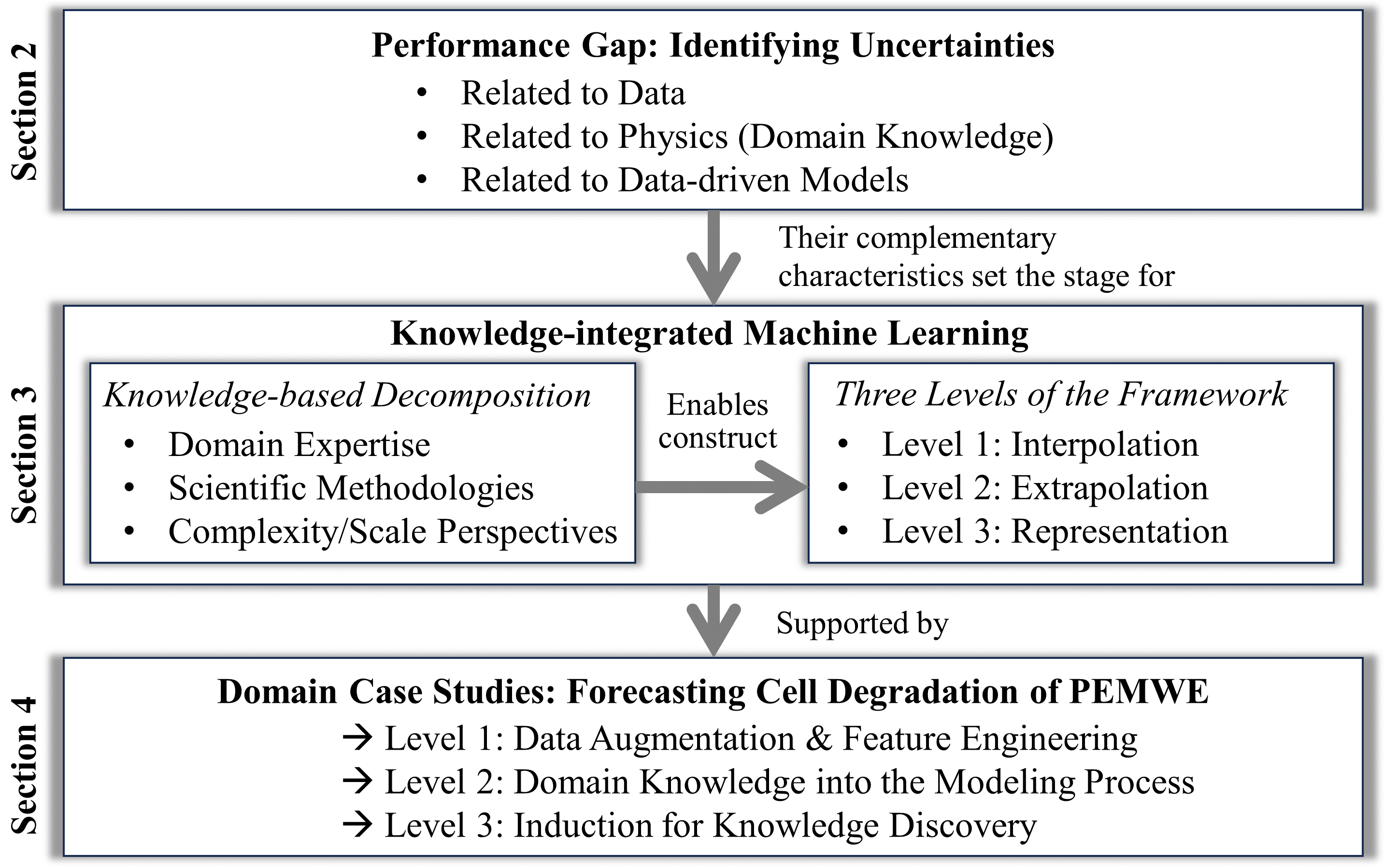}
	\caption{The manuscript skeleton of Knowledge-integrated Machine Learning in PEMWEs development}
	\label{fig:Skeleton.png}
\end{figure}

The paper is structured as follows: Within the framework of knowledge-integrated ML, Section 2 first identifies the uncertainties that originated from data, domain knowledge, and ML models, laying the groundwork for the framework introduced in Section 3, ‘The Ladder of Knowledge-integrated Machine Learning’. Furthermore, we introduced in Section 4 case studies demonstrating the framework's application in PEMWE development. Section 5 explores uncertainties and knowledge decomposition in greater depth and discusses the broader implications for the field. Finally, Section 6 summarizes our findings and outlines future research directions. Figure \ref{fig:Skeleton.png} provides a visual overview of the manuscript's structure, encapsulating the essence of our study organization.

\section{Problem Representation: Uncertainty Analysis}
\label{sec:Problem}
Recognizing the existing gap is a critical first step toward improvement. In the challenges of PEMWE development, it's essential to recognize various uncertainties that hiding behind the system description for cases such as accurate degradation forecasting and optimal system operations. These uncertainties arise from variate sources, including \textbf{Data}, \textbf{Physics (domain knowledge)}, and the inherent characteristics of \textbf{ML models} \cite{chen2023pathway} when they are involved. To systematically address these challenges, we categorize these uncertainties in PEMWEs development and examine their implications in Table \ref{tab:uncertainties}, underline their definitions, and provide relevant cases for each.

\begin{table}[p]
\centering
\caption{Categorization of Uncertainties in PEMWE Development}
\label{tab:uncertainties}
\begin{tblr}{
  width = \linewidth,
  colspec = {Q[m,105]Q[m,135]Q[m,323]Q[m,323]},
  row{1} = {c},
  cell{2}{1} = {r=3}{},
  cell{2}{2} = {c},
  cell{3}{2} = {c},
  cell{4}{2} = {c},
  cell{5}{1} = {r=4}{},
  cell{5}{2} = {c},
  cell{6}{2} = {c},
  cell{7}{2} = {c},
  cell{8}{2} = {c},
  cell{9}{1} = {r=3}{},
  cell{9}{2} = {c},
  cell{10}{2} = {c},
  cell{11}{2} = {c},
  hline{1-2,5,9,12} = {-}{},
  hline{3-4,6-8,10-11} = {2-4}{},
}
\textbf{Uncertainty Sources} & \textbf{Specific Uncertainties} & \textbf{Definition/Explanation} & \textbf{Case in PEMWE}\\
\textbf{Data} & \textit{Source Variation} & Inaccuracies from different
  origins or methodologies, causing divergent readings and challenges in
  establishing a consistent baseline or merging datasets. & Measurement discrepancies in
  PEMWEs due to varied methodologies, impacting the reliability of data
  \cite{bender2019initial, lickert2023advances, hemauer2023performance}.\\
 & \textit{System Variation} & Limited training data and
  differences between similar systems that affect model applicability and
  accuracy. & Difficulties in applying ML
  models due to a lack of sufficient data and details in system configurations
  \cite{ding2023machine, bender2019initial, gulotta2022life}.\\
 & \textit{Unbalanced Data Distribution} & Challenges in collecting and aligning data from
  various conditions evenly, and the higher presence of a certain type of system leads to data imbalance, necessitating strategies for knowledge transfer and data
  integration. & In PEMWEs, challenges in collecting fault data from diverse conditions or machinery for model training, especially with rare labeled fault data and the need for congruent datasets 
  \cite{gulotta2022life, buhre2023adaptation}.\\
\textbf{Physics (Domain Knowledge)} & \textit{Constraints on Measurability} & Challenges in accurately
  capturing complex internal processes of systems due to measurement
  limitations. & Difficulties in precisely
  measuring and interpreting aging processes of components in PEMWEs
  \cite{siegmund2021crossing, ehelebe2021limitations, lonvcar2022inter}.\\
 & \textit{Model Over-Simplification} & Simplifications in models due to
  complex nature of the system and limited cognitions in understanding the dynamics, impacting the reliability and validity of outcomes. & Simplified models fail to
  fully capture the aging processes and various PEMWE system configurations
  \cite{bensmann2022engineering}.\\
 & \textit{Confirmation Bias in
  Understanding Non-linearity and Chaos} & Cognitive biases, confirming or strengthening the beliefs or values and then challenging to dislodge once affirmed  \cite{nickerson1998confirmation}, affect the interpretation of unfamiliar phenomena from prior knowledge. & Misinterpretation of non-linear behaviors or synthetic effects in PEMWE due to preconceived notions \cite{atlam2011equivalent, bender2019initial}.\\
 & \textit{Operational Variability and
  Temporal Dynamics} & Challenges in modeling
  real-world operation variability and system temporal dynamics forecasting. & Issues in capturing the
  real-world operational variabilities and feedback mechanisms in PEMWE models
  \cite{sun2015behaviors, papakonstantinou2020degradation, feng2017review,
  suermann2019degradation}.\\
\textbf{ML Models} & \textit{Approximation Error} & Challenges in model design and
  organization, particularly in capturing complex, dynamic systems and
  processes. & Difficulties in designing ML
  models to accurately represent, align, and explain PEMWE degradation patterns, leading to "black-box" skepticism in system efficiency and operational decisions
  \cite{von2021transparency}.\\
 & \textit{Optimization Error} & Issues in finding optimal
  solutions, potentially leading
  to overcomplication or oversimplification. & Training ML models for PEMWE degradation involves balancing the model optimizing efforts from different sources of data with the practicalities of system behavior representation
  \cite{dietterich1995overfitting}.\\
 & \textit{Generalization Error} & Risks in objective function
  selection that may not align with the real-world impacts of prediction errors,
  and challenges in model over-/underfitting. & Limitations in generalizing ML
  models trained on specific PEMWE data to diverse operational conditions or
  material qualities \cite{ghosh2017robust, zhou2021asymmetric}.
\end{tblr}
\end{table}

Drawing from the insights from previous study \cite{chen2023pathway} and Table \ref{tab:uncertainties}, we observed that these uncertainty sources, though distinct, complement each other: Data uncertainties, such as those arising from measurement inconsistencies, underscore the empirical limitations present in our data collection processes. These uncertainties are further inherited by ML models, along with their mechanism approximations and assumptions, highlighting the necessity for models enriched with domain inductive knowledge, or interpretative phenomenological analysis. Meanwhile, uncertainties intrinsic to domain knowledge expose the gaps in our theoretical comprehension, necessitating the incorporation of new data and a holistic approach to reveal the system's concealed behaviors. \textbf{This synergy, where each source of uncertainty complements and informs the others, lays the groundwork for integrating domain knowledge into data-driven approaches}. 

In general, when precise and reliable modeling is crucial, addressing these uncertainties from different perspectives as much as possible is promising; this is also the case in the field of PEMWE. Integrating empirical data with domain knowledge improves model performance, where each part not only compensates for the limitations of the others but also enriches the overall accuracy and reliability of predictive models.

\section{The Ladder of Knowledge-integrated Machine Learning}
After clarifying the various sources of uncertainty and their intrinsic characteristics, we aim to bridge these insights with the existing gaps in our prior knowledge in the respective domain. To this end, this section is structured into two main segments. First, \textbf{Knowledge-based Decomposition} comprehensively maps the identified uncertainty gaps and existing knowledge frameworks. Second, the \textbf{Ladder of Knowledge-integrated Machine Learning} systematically outlines the reconciliation process between knowledge-based and data-driven modeling approaches, offering a pathway to harmonize these two.

\subsection{Knowledge-based Decomposition}
Before ascending the ladder, examining our pre-existing knowledge, especially from a decomposition perspective, is necessary. In ML approaches, the foundation for any form of learning is to capture and represent the information behind the data. This representation manifests as specific mechanisms, data structures, or other means that encapsulate knowledge effectively.

\begin{figure}[h]
	\centering
	\includegraphics[width=16.5cm]{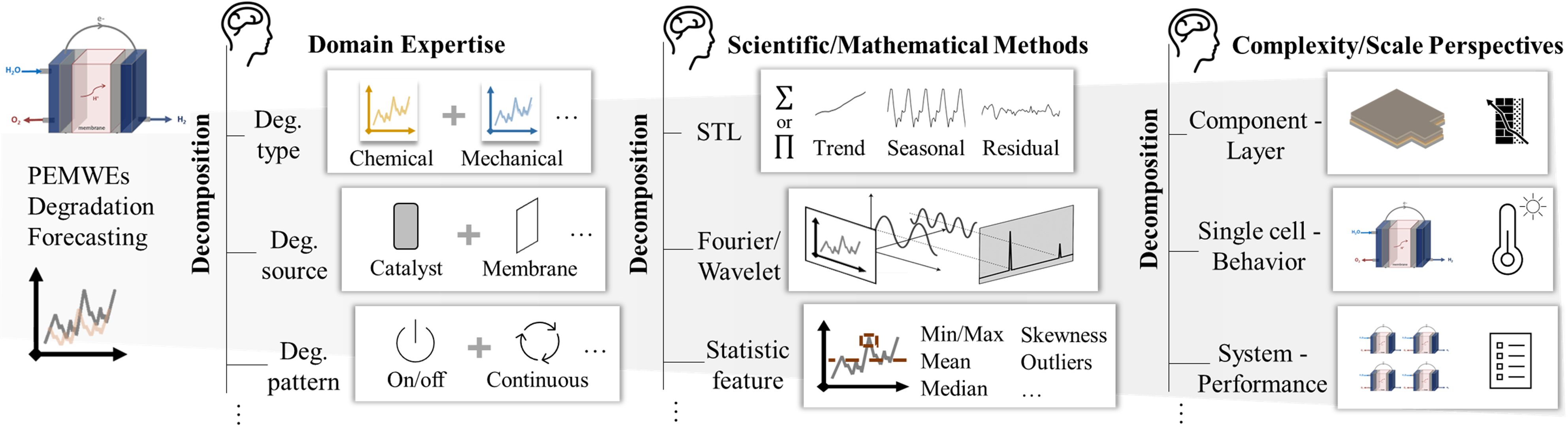}
	\caption{Three perspectives of knowledge-based decomposition in an example of degradation forecasting in PEMWE. Knowledge, including domain know-how, scientific/mathematical methodologies, and different system complexities/scales, and the exemplary respective subtopics, such as the degradation type or the single cell behavior, is embedded to help decompose the energy demand time series and gain more information to understand degradation.}
	\label{fig:Picture1.png}
\end{figure}

In engineering, many strategies exist for decomposing systems or contextualizing problems, what we call reductionism. These strategies are born from the need to understand complex systems, recognize patterns, and ultimately leverage that understanding for predictive, optimization, or decision tasks. In a complex domain like PEMWEs, many factors might influence degradation processes – from material quality and environmental conditions to operational stresses \cite{feng2017review}. We aim to isolate these variables and understand their individual and collective impacts. In this context, we provide an extensive illustration centered on the analysis of PEMWE degradation, as depicted in Figure \ref{fig:Picture1.png}, with further elaboration and exemplification in Table \ref{tab:combined_decomposition}. They serve dual purposes: to highlight the variables and interactions inherent to PEMWE systems and to present the potential of how an in-depth understanding of these elements enhances the efficacy of ML applications in engineering tasks. Specifically, the illustration identifies three principal sources of decomposition: 
\begin{itemize}
    \item \textbf{Domain Expertise}: In advancing PEMWE development, understanding and forecasting degradation is crucial for laboratory and industrial applications \cite{zio2022prognostics}. Central to this investigation is the decomposing of domain expertise. This process allows for isolating and analyzing specific factors contributing to the degradation of the system. By deconstructing domain expertise into its component parts and understanding how each influences the system, we create opportunities for a more granular problem breakdown. This approach promotes the development of sophisticated and precise models to describe the system behavior, enhancing their predictive capabilities in specific tasks.
    \item \textbf{Scientific/Mathematical Methods}: In complex systems, the data collected often carry valuable insights encoded implicitly. Scientific and mathematical processing thus emerges as an indispensable, knowledge-integrated approach in engineering endeavors. Its purpose is to distill complex data sets, extract hidden information, and eliminate noise from the observed phenomenon and data to unveil the most pertinent insights.
    \item \textbf{Complexity/Scale Perspectives}: Describing a system in a hierarchical, multi-scale manner enables us to capture its behavior across varying levels of complexity and features, thereby enhancing our understanding of the system's overall behavior. By subsequently integrating it into ML processes, we aim to construct comprehensive models that are not only data-driven but also deeply connected with the structure of domain-specific expertise.
\end{itemize}

\begin{table}[p]
\centering
\caption{Description of Knowledge-based Decomposition in PEMWE}
\label{tab:combined_decomposition}
\begin{tblr}{
  width = \linewidth,
  colspec = {Q[m,127]Q[m,127]Q[m,664]},
  row{1} = {c},
  cell{2}{1} = {r=5}{},
  cell{2}{2} = {c},
  cell{3}{2} = {c},
  cell{4}{2} = {c},
  cell{5}{2} = {c},
  cell{6}{2} = {c},
  cell{7}{1} = {r=4}{},
  cell{7}{2} = {c},
  cell{8}{2} = {c},
  cell{9}{2} = {c},
  cell{10}{2} = {c},
  cell{11}{1} = {r=4}{},
  cell{11}{2} = {c},
  cell{12}{2} = {c},
  cell{13}{2} = {c},
  cell{14}{2} = {c},
  hline{1-2,7,11,15} = {-}{},
  hline{3-6,8-10,12-14} = {2-4}{},
}
\textbf{Source} & \textbf{Sub-category} & \textbf{Examples in PEMWE context}\\
\SetCell[r=5]{m} \textbf{Domain\qquad\qquad  Expertise} & \textit{Degradation Types (Operational Stressors)} & In PEMWEs, chemical and mechanical stressors crucially affect equipment longevity and efficiency. Chemical stress, induced by impurities or mechanical stressors marked, e.g., by operational cycles, can be independently modeled \cite{chandesris2015membrane}. This separate modeling in ML methods allows for precise prediction of their individual impacts on degradation in PEMWEs.\\
& \textit{Degradation Sources (Material-Related)} & Multifaceted deterioration of PEMWE materials is analyzed for operational lifespan insights, and each aspect provides unique insights when modeled. E.g., catalyst degradation and membrane wear are studied to understand the impact of various operational conditions \cite{rakousky2016analysis}, and guide for model separation/specification.\\
& \textit{Degradation Patterns (Operational Duration) } & PEMWE longevity is closely linked to usage patterns, where domain knowledge enables distinct modeling of various operational phases. Accelerated Stress Test (AST) phases, including start-up and shut-down cycles, are crucial for assessing degradation rates \cite{weiss2019impact}. By separately modeling these AST phases and sustained continuous operations, we gain insights into their specific impacts on wear and degradation, thereby enhancing our understanding of degradation trends over different operational timescales.\\
& \textit{Operating Conditions} & Temperature variations and special operating conditions critically influence PEMWE performance and degradation \cite{frensch2019influence}. For instance, modeling temperature fluctuations separately emphasizes the need for adaptable operational strategies.\\
& \textit{Physicochemical Dynamics} & Differential equations describe the continuous changes in catalyst activity and membrane wear, providing a foundation for ML workflows to predict degradation pathways in PEMWEs.\\
\SetCell[r=4]{m} \textbf{Scientific Methodologies} 
& \textit{STL Decomposition} & Decomposition of degradation patterns into trends, seasonal fluctuations, and noise \cite{cleveland1990stl} aids ML models in adapting to layered degradation behaviors in PEMWEs.\\
& \textit{Wavelet Analysis} & Time-dependent operational signals in PEMWEs are decoded using Wavelet-analysis \cite{torrence1998practical} or Fourier-based decomposition \cite{duhamel1990fast}, identifying implicit patterns in original signals from different perspectives for data-driven models.\\
& \textit{Statistical Features as Degradation Proxies} & Basic statistical measures such as min/max voltage or accumulated number of ramps, mean values or slopes are used as proxies to understand the overall degradation landscape, guiding as indicators in the predictive model development (an example in battery aging  \cite{zhang2022machine}).\\
& \textit{Spatial Decomposition} & Analyzing from various spatial segments. For instance, identifying accelerated degradation in the center compared to outer regions due to uneven temperature and current distribution \cite{krenz2023temperature} provides critical insights. Embedding these distinct spatial aspects separately into the modeling process helps to gain a comprehensive degradation perspective, facilitating targeted interventions and system optimization strategies.\\
\SetCell[r=4]{m} \textbf{Complexity\quad Perspectives} & \textit{Component-Level Insight} & Detailed studies of individual components, such as catalyst layers and membranes, through ex-situ tests and ASTs, serve as a building block for higher-level system analysis and tailored predictive modeling.\\
& \textit{Cell-Level Understanding} & In-depth analyses of membrane health, coupled with localized environmental assessments, yield detailed insights into degradation processes within PEMWE cells. These studies, enriched by the integration of cell design variations, enable the distinct modeling of various aspects, thereby providing a granular perspective on the degradation pathways specific to each PEMWE cell.\\
& \textit{Stack-Level Perspectives} & In scaling to the stack level, segmented cell measurements and system data analysis in PEMWE stacks help predict and counteract degradation hotspots, enhancing the understanding of cell interactions.\\
& \textit{System-Level Vision} & Long-term operational protocols and peripheral equipment assessment provide a complete picture of system-level influences on degradation, forming the basis for comprehensive system modeling in PEMWEs.
\end{tblr}
\end{table}

\subsection{Three Levels of the Framework}

\begin{figure}[h]
	\centering
	\includegraphics[width=16.5cm]{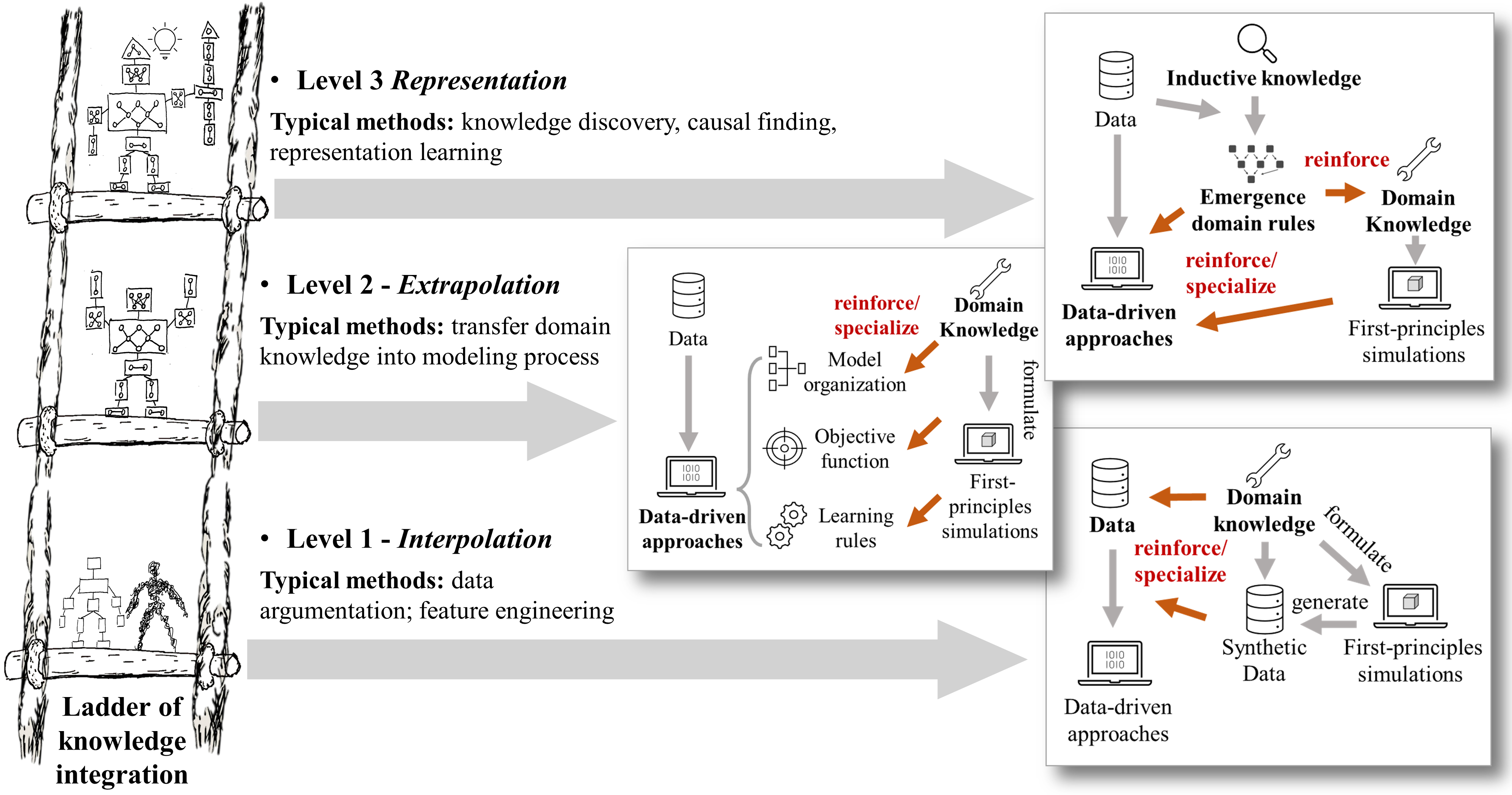}
	\caption{The Ladder of Knowledge-integrated Machine Learning \cite{chen2023pathway}. The three levels in the pathway state their difference and core ability, linking to their typical methods and characteristic descriptions. Level 1 - Interpolation: domain knowledge is embedded in data argumentation and feature engineering to achieve better performance for ML methods; Level 2 - Extrapolation: incorporating domain knowledge into the data-driven modeling process to enable informed predictions beyond the observation range of training data; Level 3 - Representation: incorporating knowledge discovery or learning mechanism into the model to transform effective information concisely. A higher level is compatible with lower abilities.}
	\label{fig:KIML.png}
\end{figure}

Once we identified the characteristics of all decomposed uncertainties regarding the gap that existed in our prior knowledge, data, and ML models, we can organize a pathway to reconcile both systematically \cite{chen2023pathway}: a three-level Ladder of Knowledge-integrated Machine Learning. This ladder combines the strengths of connectionism (data-driven techniques) and symbolism (knowledge-based methods) approaches. Each level of this ladder represents a degree of knowledge integration into the data-driven process, guiding the harmonious merger of knowledge with heuristic linkages. We name each level by its core ability:\textbf{ Interpolation (Level 1)}, \textbf{Extrapolation (Level 2)}, and \textbf{Representation (Level 3)}, as an illustrative summary presented in Figure \ref{fig:KIML.png}. Notably, in this Figure, the term of first-principles simulations offers a more generalized, high-level theoretical approach. Tailor to the PEMWE domain, we particularly focus on physics-based or empirical simulations in this study. Detailed explanations follow in the case studies.

\section{Domain Case Studies: Forecasting Cell Degradation of PEMWE}
To comprehensively present each level of knowledge integration with practical evidence, we examine specific cell degradation case studies in PEMWE by combining laboratory assessments and numerical modeling. The laboratory data consists of a dataset of almost 1500 operating hours where a single PEMWE cell was tested with a load profile consisting of repetitive characterization and step profile phases. Through voltage loss analysis using the recorded current, voltage, and Electrochemical Impedance Spectroscopy (EIS) data, the different loss terms of the total cell voltage can be calculated. The cell voltage $U_{cell}$ can be expressed as the sum of the following components:


\begin{equation}
    U_{cell} = U_{Nernst} + \eta_{act} + \eta_{ohm} + \eta_{mtx} ,
\end{equation}
where $U_{Nernst}$ represents the Nernst voltage (V), $\eta_{act}$, $\eta_{ohm}$, and $\eta_{mtx}$ correspond to activation losses (V), ohmic losses (V), and mass transport, including any remaining losses (V).
The following case studies focus on the activation losses of the cell. It should be mentioned here that the Oxygen Evolution Reaction (OER) has much slower kinetics compared to the Hydrogen Evolution Reaction (HER), and therefore, it distinctly governs this type of loss \cite{rogler2023advanced}. These losses $\eta_{act}$ can be described by the Tafel's equation:

\begin{equation}
    \label{e2}
    \eta_{act}(t) = b(t) \cdot \log_{10} \left(\frac{i(t)}{i_0(t)}\right) \ ,
\end{equation}

where $b$  is the tafel slope ($mV/dec$), $i$ is the current density ($A/cm^2$) and $i_0$ is the exchange current density ($A/cm^2$) \cite{pham2021essentials}. These parameters are depending on the time $t$.

For the level 1 case study, we use the laboratory data directly to demonstrate the accuracy improvement in a real-world scenario. Meanwhile, for the level 2 and 3 case studies, the lab data is used as the data basis for generating synthetic data to eliminate the potential biased impact from unknown hidden factors. Further information on the synthetic data follows in chapter \ref{Level 2: Extrapolation}. At all different levels, the aim is to predict the activation losses ($\eta_{act}$) as an indicator of cell degradation.


For ML approaches, we use three standard, mechanism-wise different methods in level 1 to avoid biased results due to the ML model; they are: \textit{Support Vector Regression (SVR)}, \textit{Decision Trees (DT)}, and \textit{Artificial Neural Networks (ANN)}. We modified ANNs to knowledge-integrated and applied them in level 2. In level 3, we use an optimization framework based on deep reinforcement learning called  $\phi$-SO \cite{tenachi2023deep} designed to recover analytical symbolic expressions from physics. A detailed description of each ML model and the application in this study are available in Table \ref{tab:tab2}, Appendix \ref{appendix:table}.

For performance evaluation metrics, two key metrics for model evaluation are employed: Normalized Root Mean Square Error (NRMSE) and the Coefficient of Determination (\( R^2 \)):
\begin{enumerate}
    \item NRMSE quantifies the model's prediction error. By normalizing the RMSE with respect to the range or standard deviation of the observed data, NRMSE offers a dimensionless measure that facilitates comparisons across different scales or datasets. Lower NRMSE values indicate better model performance.
    
    \item \textbf{\( R^2 \)} measures the proportion of the variance in the dependent variable that is explained by the independent variables in the model. It ranges from 0 to 1, with higher values indicating that the model explains a greater proportion of the variance. An \( R^2 \) value close to 1 suggests a good fit between the model predictions and the actual data.
\end{enumerate}

Both metrics provide complementary insights: while NRMSE focuses on the absolute fit of the model to the data, \( R^2 \) offers a relative measure of how well the model's predictions match the observed variability in the data.

\subsection{Level 1: Interpolation}
This fundamental level focuses on embedding extra information into the data inputs through pre-processing or knowledge-based methods to achieve \textbf{more accurate, robust model performance with reduced real-data reliance}. This is primarily achieved in two ways:

\begin{enumerate}
    \item \textbf{Data augmentation:} Employing first-principles simulations or numerical modeling to produce synthetic data to enrich data inputs.
    \item \textbf{Feature Engineering:} Utilizing domain knowledge to transform data to enable better machine learning performance.
\end{enumerate}

Various abstraction methods utilize first-principles methods to generate additional data \cite{chatenet2022water}. Such synthetic data acts as an extra information carrier, providing valuable supplementary patterns from our domain knowledge for the ML modeling process. This augmentation not only aids in reducing biases inherent in observational data, but also reinforces implicit patterns and introduces multi-scale data inputs. At its core, Level 1 enhances existing data with additional insights. 

Furthermore, it is possible to integrate domain knowledge with data argumentation in an incremental manner \cite{castro2018end}, where the additional knowledge input further enriches the information content of the data, tuning the input through the feedback of the model’s output, which is also referred to as feature engineering. This approach accentuates model explainability, shedding light on input feature significance and output interpretations \cite{roscher2020explainable}. For instance, \cite{chen2022hybrid} proposes a framework that simultaneously combines simulation modeling data with real-world records, streamlining the ML training process and mitigating the modeling effort to minimize performance gaps. 

In this study, we designed a comparison experiment with the earlier mentioned real laboratory data for cell degradation prediction. The goal is to predict the "observed" values from the same dataset but with/without the STL decomposition on the output. We randomly divided the dataset into an 8:2 ratio for training and testing to keep interpolation as the primary focus at this level. The \textit{time}, \textit{cell voltage}, and \textit{current density} were used as input data to train and predict the \textit{activation losses} with three different ML architectures (SVR, DT, and ANN). Please refer to Appendix \ref{appendix:l1} for their modeling details.

\begin{figure}[h]
	\centering
	\includegraphics[width=16cm]{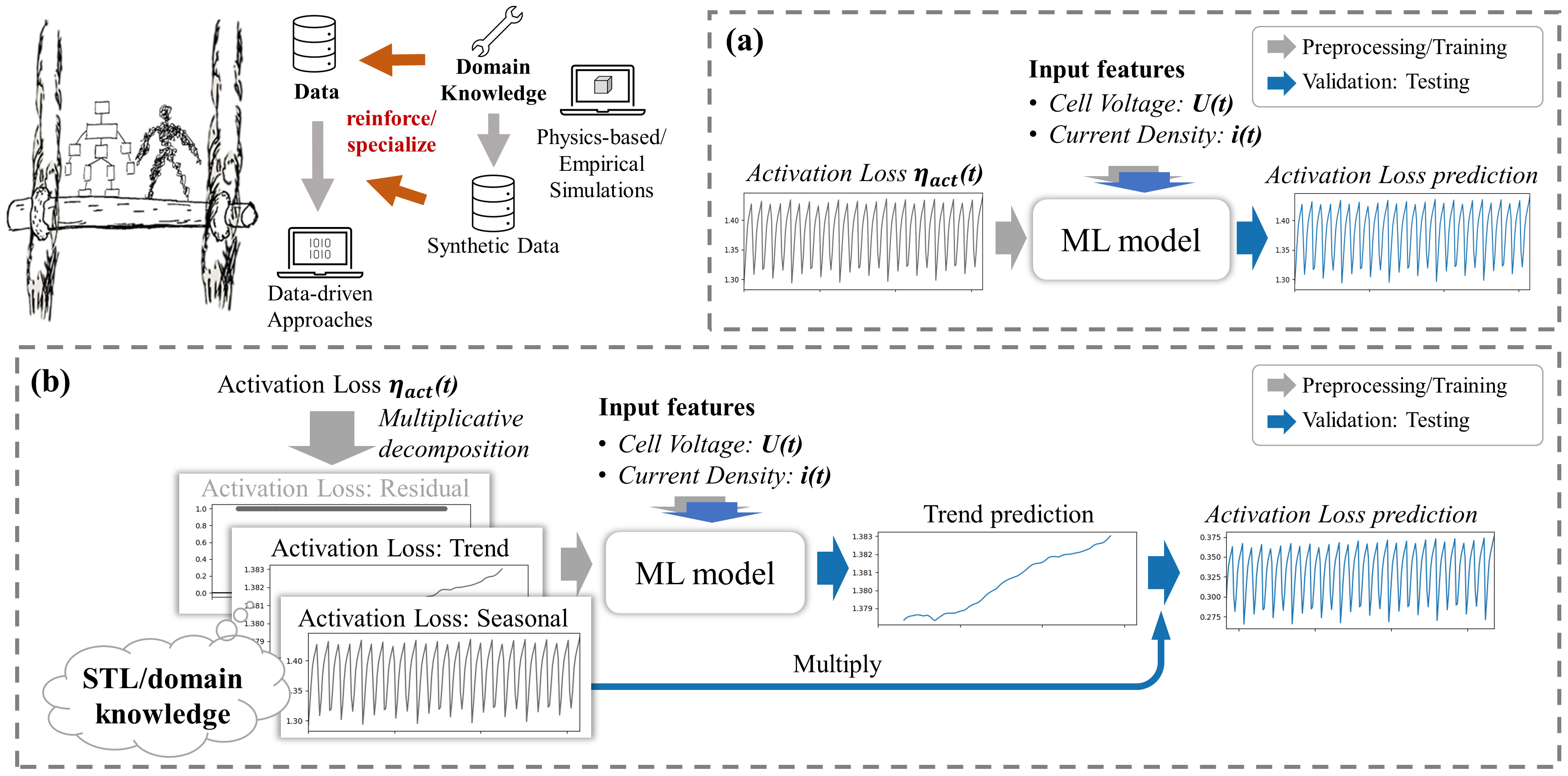}
	\caption{Level 1 case: STL Decomposition of the activation loss curve for the test set prediction. (a) Standard ML prediction: ML model directly predicts the activation losses using \textit{time duration}, \textit{cell voltage} and \textit{current density} as input features; (b) Knowledge-integrated ML prediction (STL without residual): activation losses are decomposed into multiplicative trend, seasonal, and residual components using prior knowledge. The ML model trains on and predicts the trend, which is then multiplied by the known seasonal pattern.}
	\label{fig:L1_case.png}
\end{figure}

With the STL decomposition, we use the same seasonal cycling period for both training and testing datasets to decompose the \textit{activation losses} curve into trend, seasonal, and residual patterns, training ML models to fit and predict only the trend curve on the testing set, conclude the result by reattaching the seasonal pattern into the prediction. Comparisons were drawn between standard ML predictions of \textit{activation losses} (subfigure (a) in Figure \ref{fig:L1_case.png}) and predictions derived from the product of decomposed loss trend with known seasonal patterns (subfigure (b) in Figure \ref{fig:L1_case.png}). Compared to the results without STL decomposition, we observed a general accuracy improvement on the test set across different ML models, as shown in Table \ref{tab:tab3}. 

\begin{table}[h]
\centering
\caption{Results from the case in level 1. Three different mechanism-wise ML approaches are involved. With generally lower $NRMSE$ and higher $R^2$ results, the prediction accuracy with embedded STL decomposition (right) is higher than standard ML prediction (left).}
\begin{tblr}{
  width = \linewidth,
  colspec = {Q[92]Q[158]Q[133]Q[271]Q[260]},
  column{even} = {c},
  column{3} = {c},
  column{5} = {c},
  cell{1}{2} = {c=2}{0.291\linewidth},
  cell{1}{4} = {c=2}{0.531\linewidth},
  hline{1-3,6} = {-}{},
}
 & \textbf{\textit{ML prediction}} &  & \textbf{\textit{ Knowledge-integrated ML prediction (STL without residual)}} & \\
 & \textbf{$NRMSE$} & \textbf{$R^2$} & \textbf{$NRMSE$} & \textbf{$R^2$}\\
SVR & 39.883 & 0.832 & \textbf{4.996} & \textbf{0.971}\\
DT & 8.470 & 0.916 & \textbf{4.315} & \textbf{0.978}\\
ANN & 38.340 & 0.824 & \textbf{3.238} & \textbf{0.988}
\end{tblr}
\label{tab:tab3}
\end{table}

The result from Table \ref{tab:tab3} supports that integrating domain knowledge, even a simple decomposition engineering to remove the known seasonal patterns from data, effectively improves data-driven models' accuracy. 

However, it should be noted that while level 1 mitigates data reliance issues to some extent and is more tailored for specialized domains, it doesn't alter the intrinsic efficiency of the data-driven method. Despite its blended nature, the model retains an interpolative behavior in high-dimensional spaces \cite{smith2014towards}. In other words, although the knowledge integration at this level improves the model performance, the model's reliability remains constrained to the bounds of the input data, and the model itself remains the black-box model characteristics.

\subsection{Level 2: Extrapolation} \label{Level 2: Extrapolation}

At level 2, the focus shifts from merely augmenting data itself (as seen in level 1) to embedding domain knowledge into the design of data-driven models, which enables the model to conduct plausible and better outcomes beyond the original in the specific domain. This could be achieved from various perspectives in the modeling process:

\begin{enumerate}
    \item \textbf{Objective Function: }By defining what constitutes a quality prediction, domain knowledge forms a nexus between data and model architecture. For example, the Physics-Informed Neural Network (PINN) melds partial differential equations (PDEs) into the neural network's loss function \cite{raissi2019physics}, allowing it to effectively emulate physical processes like heat transfer \cite{drgovna2021physics}.
    \item \textbf{Learning Rule: }Knowledge integration introduces constraints and regularizations to the learning algorithms. This optimizes the learning efficiency and ensures the validity of outcomes. The Bayesian Physics-Informed Neural Network (B-PINN) stands out here, incorporating a Bayesian framework to the traditional PINN, making learning more resistant to noise and model perturbations \cite{yang2021b}.
    \item \textbf{Model Organization: }This encompasses restructuring the model by integrating domain-specific knowledge structures. We refer to an instance in the building engineering domain: using a Component-Based Machine Learning (CBML) approach that segments system structures and trains ML models at discrete scales (components, zones, buildings) to consist in fitting new buildings. Previous research shows this segmented methodology contributes to the modeling flexibly, enhancing their accuracy and utility efficiency in the domain \cite{geyer2018component, chen2023utilizing}.
\end{enumerate}
Conclusively, Level 2 is distinguished by its in-depth tailor of knowledge representation into the data-driven modeling and fitting process, creating \textbf{more domain-specific, interpretable models and capable of extrapolation}. Essentially, it embeds the model process with the logistic structure of prior knowledge and enables the model to conduct informed predictions outside of observable ranges.

At this level's case study, we designed a comparison experiment between ANN and PINN in forecasting the activation losses of a PEMWE cell. For this, we elevated the experiment by harnessing generated data, armed with the precise formula (prior knowledge), to forecast the \textit{activation losses}. Based on Tafel's equation to generate the synthetic dataset, as described in \cite{pham2021essentials} and Equation \ref{e2}, we introduced Gaussian noise denoted as $\epsilon$ with a standard deviation of one. This noise addition served to simulate the operational conditions realistically:
\begin{equation}
\eta_{act}(t) = b(t) \cdot \log_{10}\left(\frac{i(t)}{i_0(t)}\right) + \epsilon \ .
\label{eq:uloss}
\end{equation}
In this equation, the time-varying \textit{Tafel Slope} $b$ and \textit{Exchange Current Density} $i_0$ were individually approximated using fifth-degree polynomials derived from experimental data. With the aid of these polynomial expressions, we generated a synthetic dataset to simulate 2800 hours of operation.

\begin{figure}[h]
	\centering
	\includegraphics[width=16.5cm]{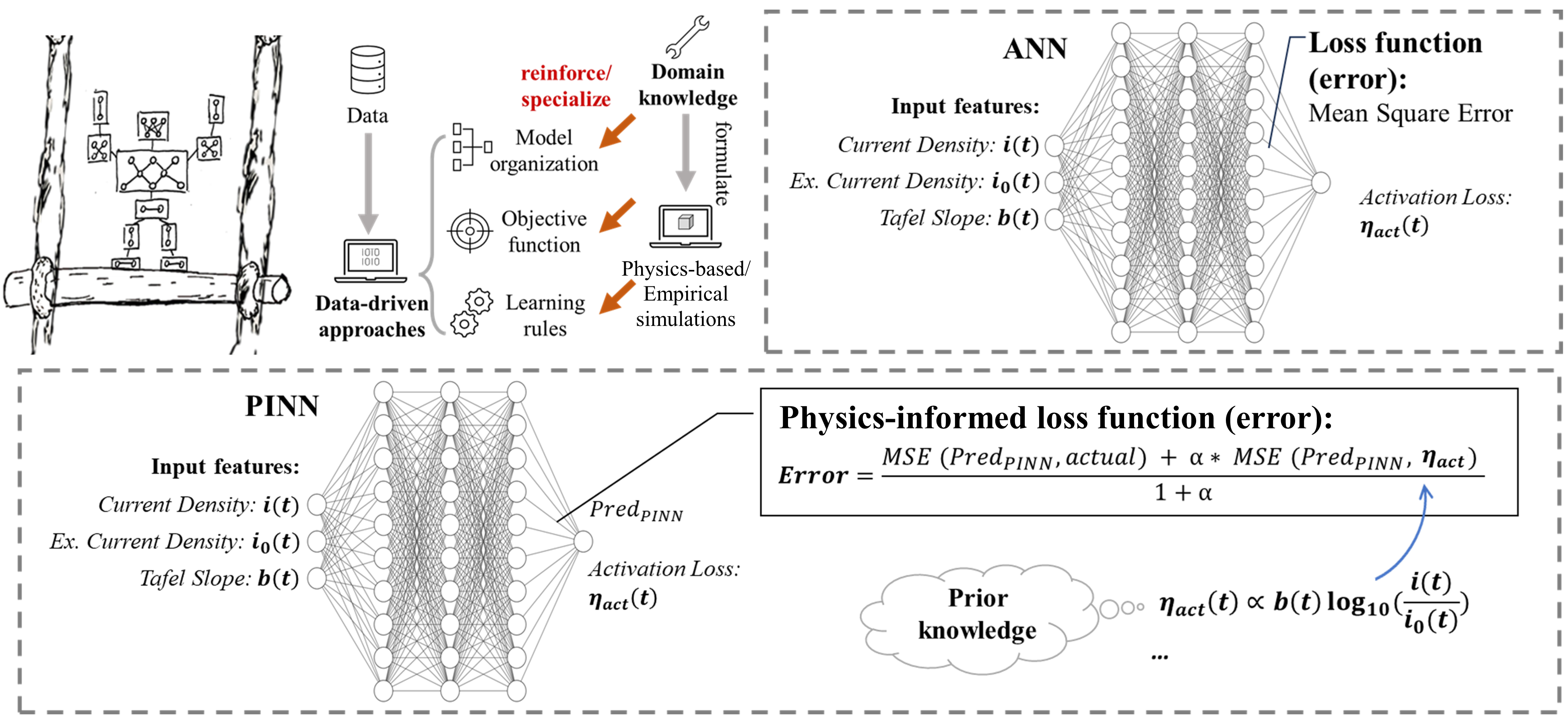}
	\caption{Level 2 Case: Physical-Informed Neural Network for activation loss forecasting with partial knowledge. (a) Vanilla ANN organization. (b) Physical-Informed Network Organization: Integrates embedded prior knowledge and consists of two networks. The physical-informed network with a loss function comparing prior knowledge results and prediction outcomes. The outcomes of both models are combined using weighted sums to produce the final training output, where $\alpha$ adjusts the output weight between the PINN and Correction Model in the loss function.}
\label{fig:L2_case.png}
\end{figure}

In this case, we presented a PINN specifically designed for \textit{activation loss} forecasting, seamlessly blending data-driven insights with select domain expertise. A comprehensive visualization of the level 2 case is provided in Figure \ref{fig:L2_case.png}. We used a Vanilla ANN with a mean square error (MSE) loss function for benchmarking. Architecturally, it features an input layer consisting of three neurons, encompassing inputs of \textit{Tafel Slope} $b$, \textit{Current Density} $i$, \textit{Exchange Current Density} $i_0$, and, succeeded by three hidden layers with ten neurons each, and culminates in a single-neuron output layer. This specific design ensures a balanced architecture, optimizing learning and prediction efficiency in alignment with our research goals. The data was sequentially divided into an 8:2 ratio for training and testing, transitioning the study from interpolation to extrapolation.

The PINN is grounded in physical principles and employs a loss function that compares MSE values derived from prior knowledge predictions and model outcomes. Concurrently, mirroring the Vanilla ANN mentioned earlier, it counters potential biases by incorporating prior knowledge. The general organization adeptly extracts patterns from empirical data and integrated domain expertise, ensuring its predictions are precise and devoid of undue biases. The PINN is trained with a unique loss function by taking the MSE between the prediction outcome of the $Pred_{PINN}$ and the actual value $\eta_{act}$, calculated by domain knowledge, as well as the MSE between the $Pred_{PINN}$ and the ground-truth value $actual$ into consideration, as described in the following equation:

\begin{equation}
Error = \frac{MSE(Pred_{PINN},actual) + \alpha*MSE(Pred_{PINN}, \eta_{act})}{1+\alpha}
\label{eq:error}
\end{equation}

For a comparative study in different configurations, we varied the weight $\alpha$ in the loss function of the PINN between 0.1 and 5, and conducted various partial knowledge integration, including combinations of different variables and operators from the Equation \ref{eq:uloss}. Each model runs under different configurations for 20 rounds to conclude the result statistics. The result is summarized and visualized in Figure \ref{fig:L2_result.png}, as more modeling detail is available in Appendix \ref{appendix:l2}.

\begin{figure}[h]
	\centering
	\includegraphics[width=16.5cm]{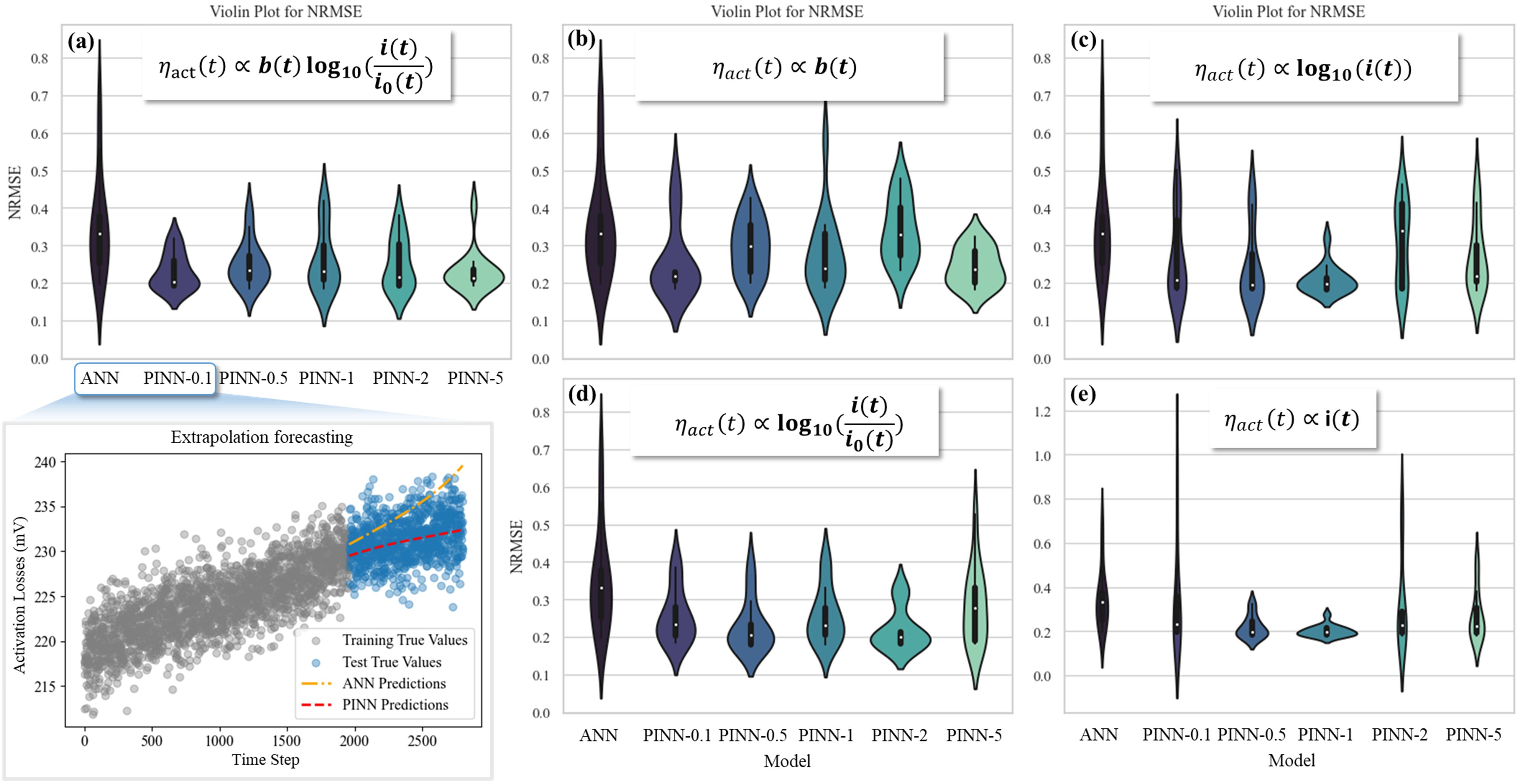}
	\caption{Level 2 case result: Model performance in violin plot, NRMSE. Each model undergoes 20 rounds with each configuration, with results shown in the violin plot. The y-axis shows the model performance in NRMSE, while the x-axis indicates different models, e.g., PINN-0.1 means PINN organization with weight $\alpha$ set as 0.1. Sub-graphs (a)-(e) depict varying degrees of prior knowledge integration within the PINN, as the corresponding prior knowledge indicated in each subplot's mid-top position, compared with a pure ANN. Specifically, sub-graph (a) illustrates extrapolation forecasting contrasts between ANN and PINN.}
\label{fig:L2_result.png}
\end{figure}

As shown in Figure \ref{fig:L2_result.png}, PINNs outperform the ANN model in most configurations. Even with partial knowledge (equation) integration (sub-graph (b)-(d)), there's a notable improvement in prediction accuracy, suggesting that the careful inclusion of domain knowledge contributes to accuracy improvement and helps reduce performance fluctuations. Notably, sub-graph (a) offers a more detailed view of this performance disparity in the context of extrapolation forecasting. The scatter plot contrasts the predictions of ANN and PINN against the true values of the synthetic dataset. This superiority is particularly evident for PINN-0.5 and PINN-1. Both show improved robustness with narrower NRMSE distributions. The distinct trajectories of ANN and PINN predictions highlight the latter's better forecasting ability, especially in the given extrapolation scenario. PINN's predictions adhere more closely to the extrapolated ground-truth values in the testing set, showcasing its robustness benefit by the extra embedded domain knowledge.

A closer look at sub-figure (e) in Figure \ref{fig:L2_result.png} reveals a notable issue: the omission of the log operation relationship in the prior knowledge leads to a deterioration in performance compared to other configurations. A potential reason for this could be that the role of logarithmic transformation is vital in stabilizing variance and provides the model with effective information. In its absence, the model struggles to correctly capture the complexity of the underlying relationships, thereby compromising the forecast accuracy.

Furthermore, it's worth noting that variations in the weight $\alpha$ within different PINN configurations seem to have minimal impact on the overall performance. This underscores the resilience of the PINN framework, suggesting that its efficacy isn't overly reliant on precise weight adjustments.

In conclusion, the case study result supports the proposed ladder's level 2 effectiveness. Strategic integration of prior domain knowledge significantly improves forecasting accuracy and demonstrates that pure ML methods cannot perform extrapolation tasks. Yet, the degree of this integration must be chosen carefully, as an excessive or minimal integration might not yield the desired improvements.

\subsection{Level 3: Representation}

Level 3 signifies a status transition from direct domain knowledge integration to a more profound understanding of inductive principles. As the major effort in the previous two levels lies in integrating known prior knowledge into the modeling process to teach ML models 'learn to do', the objective at this level is broader: to enable models to \textbf{discover} \cite{lake2017building} \textbf{the implicit, representative patterns directly and concisely from data without potentially biased prior knowledge input.} This entails:

\begin{enumerate}
    \item \textbf{Autonomous Knowledge Extraction:} Instead of relying on direct known, domain-specific knowledge input, the ML processes autonomously discern and address engineering problems in an end-to-end manner, translating them into representations that are either unsupervised \cite{khosla2020supervised} or semi-supervised \cite{jing2022masked, van2020survey}.
    \item \textbf{Shift to Second-order Knowledge:} This level prioritizes knowledge that enables the modeling process to discover and generalize transferable patterns autonomously. By incorporating this second-order knowledge, models can discern domain rules without supervision, thus reducing biases from prior knowledge.
\end{enumerate}
Examples of second-order knowledge techniques include:

\begin{itemize}
    \item \textit{Causal Dependencies Asymmetry}: Employing causal inference techniques \cite{pearl2000models} to determine not just correlations but causal relationships between variables, enabling the model to answer in 'what-if' scenarios for similar cases. For instance, the extracted causal skeleton from variables is transferable into new cases and benefits the efficiency of design and engineering scenarios. We point to successful cases in the building engineering domain \cite{chen2022introducing, chen2024using}.
    \item \textit{Entropy Reduction}: Techniques like self-supervised learning focus on optimizing representation quality by discriminating between essential and non-essential information, and formulating it into a more compressed representation. Despite the high-dimensional manifold learning \cite{lin2008riemannian} or semi-supervised learning \cite{jing2022masked,khosla2020supervised,van2020survey} that is hard to interpret by prior knowledge. One of the possible formats in our domain could be symbolic optimization for knowledge discovery \cite{tenachi2023deep}.  
\end{itemize}

At this stage, our methodology focuses on the symbolic optimization for knowledge discovery. Utilizing the same dataset from level 2, level 3 adopts an exploratory strategy to directly extract the encapsulated formula (Equation \ref{eq:uloss}) from the data. To achieve this, we adapted Physical Symbolic Optimization ($\phi$-SO) \cite{tenachi2023deep}, rooted in deep reinforcement learning techniques. These methods aim to obtain an analytical formula approximation from noisy data.

It's important to note that the prior knowledge at this level isn't the formula itself, but rather the symbolic conceptual abstraction of physical units and operators. These operators include addition, subtraction, multiplication, exponentiation, logarithm, etc. To ensure the physical relevance and consistency of the solutions presented by $\phi$-SO, every defined variable is allocated its respective International System of Units (SI). This assignment not only guarantees the variable's meaning in physics but also provides essential constraints for the optimization process, ensuring that the proposed solutions remain physically coherent.

\begin{figure}[h]
	\centering
	\includegraphics[width=16.5cm]{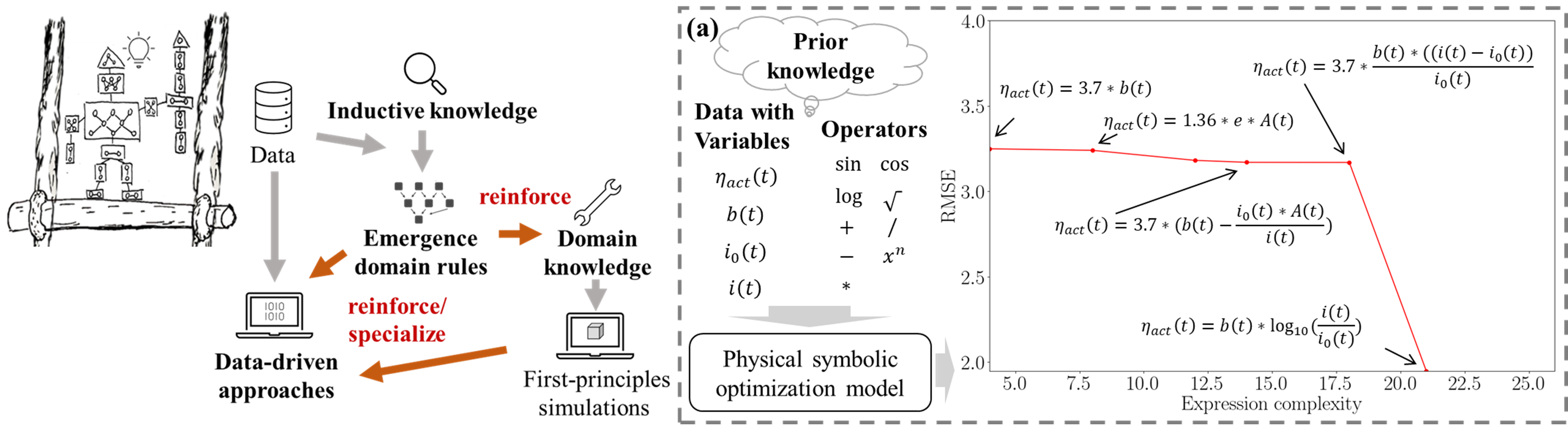}
	\caption{Level 3 Case: Integrating knowledge discovery mechanism into the model. (a) symbolic optimization: the model identifies the optimal formula combination for domain data fitting using predefined variables with physical units and operators. $\phi$-SO \cite{tenachi2023deep} is used in this case for optimization.
}
\label{fig:L3_case.png}
\end{figure}

This approach emphasizes the 'meta-knowledge' of defined operators integrated with constraints on physical units to discover domain-specific formulas from data directly. The discovery process within this framework is propelled by deep reinforcement learning, which employs unit constraints and an optimization strategy focused on minimizing prediction errors. This method enables the identification and extraction of physically coherent formulas present in the data, highlighting the primary skill at this level: Representation. Figure \ref{fig:L3_case.png} depicts the knowledge discovery process. By evaluating the performance of the retrieved symbolic expressions during the training process, the model ultimately determines the best-fit formula, as shown in Equation \ref{eq:uloss} after several runs. Further details are available in Appendix \ref{appendix:l3}.

Finally, it's important to clarify that level 3 is not a direct extension of level 2's domain knowledge. Specifically, in our case, such 'meta-knowledge' that supports modeling directly from the data plays a pivotal role in shaping the model for knowledge discovery: The operator rules we define serve as constraints on the learning algorithm, while the integration of symbolic regression tasks shapes the model's objective function. Additionally, the adoption of a deep reinforcement learning paradigm informs the model's organizational structure, enabling it to generate both accurate and physically meaningful symbolic expressions. 

While this knowledge discovery method is domain-agnostic, the integration of domain knowledge, as presented in Level 2, seems to be bypassed in this context. However, the knowledge discovered at level 3 can subsequently be utilized as 'domain knowledge' in level 2's paradigm for specific tasks, such as forecasting. This interplay between levels 2 and 3 helps to create a data-driven pipeline to foster a closer integration between data and knowledge by automated bridging between both sides. Moreover, the principles underlying Level 1 can always be beneficial and leveraged to enrich the information from data, thereby facilitating more efficient model training.

\section{Discussion}
\label{sec:Discussion}

While traditional data-driven algorithms are crucial for model construction based on identifiable patterns, this study demonstrates that incorporating empirical knowledge significantly enhances model performance, plausibility, transferability, and the overall utility of augmented intelligence \cite{ashby1956introduction}. Here are some further thoughts to contribute to this topic:

\subsection{Uncertainties with Knowledge Alignment}

Complex engineering systems, exemplified by PEMWE, inherently come with multifaceted uncertainties. These uncertainties can emerge from various fronts, such as materials used, operational conditions, or inherent non-linearities in the system dynamics. Understanding and accurately representing these uncertainties is paramount for developing reliable and robust models. A knowledge-integrated approach offers a comprehensive way to address these challenges, yielding several noteworthy benefits:

\begin{enumerate}
    \item \textbf{Unified Understanding with Quantified Uncertainty:} While most ML methodologies excel in identifying patterns within large datasets, their effectiveness is compromised by noise or data variations, highlighting the importance of detailed uncertainty analysis. Such analysis could benefit significantly from domain knowledge interpretation for further improvements. For instance, when making predictions about the voltage efficiency in the electrolysis process, domain expertise can pinpoint the sources of discrepancies, whether they stem from material inefficiencies, temperature fluctuations, or pressure deviations in the cell, and decompose the data/model accordingly. Such detailed understanding not only ensures that actual data inform predictions but is also grounded in domain scientific principles. This leads to a more quantified comprehension of uncertainty sources. Furthermore, integrating data and data-driven prediction methods helps substantiate the range of uncertainty in complex systems, thereby contributing to new understandings in the domain. This creates a dynamic feedback loop, accelerating the development through a continuous data collection cycle, model refinement, knowledge enrichment, and data re-evaluation.
    \item \textbf{Interpretable Performance:} Traditional loss functions, like Mean Squared Error (MSE), essentially quantify the difference between predicted and observed values. Yet, in complex systems, this might not be adequate. The introduction of a custom loss function, which integrates MSE with a physics-informed component, is a testament to this realization. For example, consider the degradation forecasting in PEMWE. Degradation often results from many factors like carbon corrosion, membrane thinning, or catalyst oxidation. A purely data-driven model might misguide an engineer to attribute a sudden change in performance to one factor, overlooking others. However, a loss function design based on physics ground can provide weighted importance to these factors based on prior knowledge, ensuring that the model's predictions do not only rely entirely on empirical data but also resonate with the underlying physical processes. 
\end{enumerate}

\subsection{Broader Implications for Engineering}
The mindset of comprehensively embedding domain knowledge into ML models not only promises advancements in PEMWE but also offers noticeable benefits for a wide range of engineering domains. This approach's implications can be summarized as follows:
\begin{enumerate}
    \item \textbf{Enhanced Generalizability Across Varied Conditions:} Complex systems operate in a vast parameter space. Traditional ML models, when trained on limited datasets, risk overfitting and losing efficacy in new or varying conditions. By integrating a knowledge-based approach that incorporates principles of thermodynamics, kinetics, or material science into the model's structure (e.g., component-based ML), we can create hybrid methodologies. These approaches present the potential to generalize more effectively across a broader spectrum of conditions, demonstrating more reliable behavior than models relying solely on data-driven techniques \cite{chen2022hybrid, geyer2018component, chen2021component}.
    
    \item \textbf{Robustness in Scaling Challenges:} Technology developments often transit from laboratory prototypes to commercial scales, where system dynamics may evolve or become more complex. A model built upon fundamental principles with different scales makes adapting to these scaling changes more flexible than just fitting to a specific dataset. This robustness becomes especially critical in predictive maintenance or long-term performance assessments, where the stakes are high in industrial applications.
    
    \item \textbf{Transferability and Cross-Disciplinary Applications:} The framework we developed for integrating domain-specific knowledge into data-driven models is not solely confined to PEMWE; it is adaptable to various similar and even disparate technologies. While the specific kinetics and dynamics might differ, the foundational methodological frameworks—like physics-informed loss functions or thermodynamic constraints—can be universally applied. This extends the technology's reach beyond electrochemical systems and offers a unified framework for other engineering disciplines. For instance, these principles of knowledge integration can be beneficially applied to other complex systems, such as renewable energy grids, wastewater treatment processes, and advanced manufacturing setups. In each of these areas, the same balanced approach between data-driven and physics-based modeling can result in significant gains in efficiency, robustness, and sustainability.
\end{enumerate}

\subsection{Knowledge Discovery}

Finally, in advancing the concept of knowledge discovery within configuration at level 3, as it is set at the top in the hierarchy of knowledge-integrated ML, this level is more than a sophisticated algorithmic playground; it offers a holistic strategy for addressing challenges often encountered during model training:
\begin{enumerate}
    \item \textbf{Symbolic Operations and Implications}: The symbolic operations in level 3 sets the stage for a delicate approach to equation generation. For instance, consider a scenario in PEMWE where understanding the relationship between temperature, pressure, and reaction rate is pivotal. Traditional ML models might predict outcomes based on past data but in a black-box manner. Level 3, with its symbolic operations, offers a more transparent and dynamic approach. For instance, consider the scenario where the tafel slope, a critical aspect in understanding electrochemical systems, is not previously known. In such a case, Level 3 could play a pivotal role in explaining the discrepancies observed in the predictions at Level 2 by dynamically generating and refining equations based on data, and potentially revealing new relationships or patterns. Such discoveries offer actionable insights to engineers grounded in a deeper understanding of the physical processes underlying the data.
    \item \textbf{Iterative Knowledge Discovery and Continuous Evolution}: Level 3's knowledge discovery is not a one-off event, but rather a continuous process where scientists revisit and refine hypotheses based on new data. Delving into the PEMWE example, as more data on electrode degradation over time is accumulated, level 3's symbolic modeling can adapt. It might initially have modeled degradation linearly, but with more data feeding in, it could discern and adapt to a nonlinear, perhaps exponential, decay pattern. This iterative refinement offers a two-fold benefit: It allows for direct comparison of discovered knowledge with the state-of-the-art body of knowledge, either providing valuable confirmation in case of matching or an incentive to develop this body further in case of deviation. In the latter case, it fine-tunes predictions and reveals previously hidden insights into the PEMWE process, thereby enriching our understanding from another perspective at different phases.
    \item \textbf{Cybernetic Perspective and Augmented Intelligence}: The dynamism of Level 3's approach evokes concepts of a cybernetic system—a system that observes, learns, and self-corrects. The cyclic feedback mechanism of knowledge integration discussed before exemplifies how insights from level 3 enrich the foundational methodologies of level 2 and level 1. In this study, a revelation about the interaction between PEMWE's membrane components discovered at level 3 could be integrated into level 2's modeling techniques, enabling a feedback loop for a more comprehensive understanding. This dynamic, interconnected knowledge system extends ML from mere computational tools to an embodiment of augmented intelligence, symbiotically co-evolving with human expertise.
\end{enumerate}

\section{Conclusion}
\label{sec:Conclusion}
In this study, we proposed a systematic framework for integrating domain knowledge into machine learning techniques, particularly in addressing uncertainties within complex systems like Proton Exchange Membrane Water Electrolyzers (PEMWEs). This approach aims to take a step forward toward the application of AI for science and strives to strike a delicate balance between empirical accuracy and physical plausibility. The proposed methods at different levels, combining the strengths of data-driven techniques with in-depth domain expertise, offer a profound potential to unveil methodologically sound insights that are deeply aligned with real-world applications. The proposed framework, merging empirical data with robust physics-based principles, is poised to accelerate not only the PEMWE system developments but also pave innovation paths exploiting and integrating both knowledge-based reasoning and data-driven techniques across the engineering landscape.

\section{Acknowledgement}
\label{sec:Acknowledgement}
The authors gratefully acknowledge the financial support by the Federal Ministry of Education and Research, Germany, in the framework of HyThroughGen (project number 03HY108C), DERIEL (project number 03HY122G) and SEGIWA (project number 03HY121G). Furthermore, the authors acknowledge the support of Deutsche Forschungsgemeinschaft (DFG) by funding the research through the grant within the researcher unit FOR2363 EarlyBIM (GE1652/3-2) and the Heisenberg grant for Philipp Geyer (GE1652/4-1). 

\section{Appendix}
\appendix
\label{sec:Appendix}

\section{Data-driven Methods}
\label{appendix:table}
All models used in this study, with their basic mechanism and their application scenario, are summarized in Table \ref{tab:tab2}. 

\begin{table}[h]
\centering
\caption{Data-driven methods used in the case study}
\begin{tblr}{
  width = \linewidth,
  colspec = {Q[100]Q[482]Q[351]},
  hline{1-6} = {-}{},
}
\textbf{Model} & \textbf{Description} & \textbf{Application in the study}\\
\textbf{Support Vector Regression (SVR)} & A type of Support Vector Machine (SVM) used for regression tasks \cite{drucker1996support}. The primary objective of SVR is to find a function that has at most $\epsilon$ deviation from the actually obtained targets for all the training data, and at the same time, is as flat as possible. In simple terms, SVR aims to find the best line (or hyperplane in higher dimensions) that fits the data with minimal error and within a predefined margin. & In level 1, SVR is employed to predict the \textit{Activation Loss} using laboratory
  data, benefiting from its capability to handle non-linear relationships and
  outliers efficiently.\\
\textbf{Decision Trees} & A type of flowchart-like structure in which each internal node represents a 'test' on an attribute, each branch represents the outcome of the test, and each leaf node represents a class label (or a continuous value in the context of regression) \cite{quinlan1986induction}. The paths from the root to the leaf represent decision rules, which makes it simple to understand and interpret as a
  data-driven method. & Utilized in Level 1, Decision Trees offer a transparent and interpretable approach to predict \textit{Activation Loss}, allowing for an intuitive understanding of the underlying patterns in the data.\\
\textbf{Artificial Neural Networks (ANN)} & Computing systems inspired by the biological neural networks that constitute animal brains \cite{mcculloch1943logical}. These systems learn (progressively improve their performance) to do tasks by considering examples. An ANN comprises layers of interconnected nodes (or 'neurons'). Each connection has a weight, which is adjusted during learning according to a learning rule. The strength of the weight indicates the importance of the particular input to the neuron. & In Level 1, ANNs are used alongside other models to predict \textit{Activation Loss}. Their flexible and adaptive nature makes them apt for capturing complex non-linear relationships in the data. 
  
In level 2, an ANN with a specific structure is constructed to integrate domain knowledge into the prediction task, leveraging its vast representational power.\\
\textbf{Physical Symbolic Optimization ($\phi$-SO)} & A Physical Symbolic Optimization framework designed to recover analytical symbolic expressions from physics data using deep reinforcement learning techniques \cite{tenachi2023deep}. It emphasizes ensuring the physical units consistency by construction, enabling a more focused search and improved performance. & In level 3, $\phi$-SO is employed for knowledge discovery, aiming to autonomously rediscover the formula for \textit{Activation Losses} using the dataset, proving its efficacy in uncovering underlying physical relationships from noisy data.
\end{tblr}
\label{tab:tab2}
\end{table}

\section{Comparison Experiment Description - Level 1}
\label{appendix:l1}

This comparative experiment aims to predict the 'observed' values from the same dataset but with/without the STL decomposition on the output. Data normalization is performed in input features using the StandardScaler to ensure model stability across the three chosen models, mainly from the scikit-learn package \cite{pedregosa2011scikit} in Python: SVR, Decision Tree, and ANN. Their configurations are described as follows:

\begin{itemize}
    \item \textit{Support Vector Regression (SVR)}: Initialized with default hyperparameter setting, the SVR model is trained on the scaled data. Post-training, predictions are inverse-transformed to the original data scale for subsequent performance assessment.
    \item \textit{Decision Tree Regressor}: The model is trained on the unscaled dataset, given that the splitting mechanism of tree-based algorithms is not sensitive to data scale. A default hyperparameter setting is utilized for its configuration.
    \item \textit{Artificial Neural Network (ANN)}: The ANN model is specified with two hidden layers, each containing 50 neurons ($hidden\_layer\_sizes=(50, 50)$), and an iteration limit set to 1000 ($max\_iter=1000$). The model is trained using scaled data, and predictions are then inverse-transformed to match the original scale.
\end{itemize}

In a comparative experiment, the focus shifts from predicting the 'observed' values to predicting the 'trend' component from a decomposed dataset. Following similar data preparation and normalization steps, the same three models are trained. Evaluations in this experiment focus on the model predictions with respect to their seasonal and residual components, both with and without the residual's influence. The final evaluation phase computes the \textbf{normalized root mean squared error (NRMSE)} and \textbf{Coefficient of determination ($R^2$)} for each model to measure their predictive performance against actual observations.

This consolidated framework offers a comprehensive evaluation of three different machine learning models across two prediction scenarios, enabling a robust assessment of their predictive capabilities within the given dataset.

\section{Comparison Experiment Description - Level 2}
\label{appendix:l2}

In this case, we designed a dual-element Physics-Informed Neural Network (PINN). The purpose of the following content is to elaborate on the modeling detail not already addressed in the manuscript.

\textbf{Neural Network Architecture and Data Preparation}

Leveraging the PyTorch \cite{paszke2019pytorch} package in Python, we implemented a feed-forward Artificial Neural Network (ANN) having three hidden layers. The neural architecture begins with an input layer of three neurons, progressing through three hidden layers containing ten neurons each, and culminating in an output layer. Non-linear activations in the form of ReLU functions are interspersed between these layers. PINN organization comprises two architecture-wise identical ANNs; one serves as the correction model, and the other uses a customized loss function to be physics-informed, as the true value is calculated based on available inputs and prior knowledge.  

Initial data is synthetic with a standard deviation of one Gaussian noise added. Input feature standardization is achieved via the StandardScaler from scikit-learn. The dataset is then divided into training and test sets and converted into PyTorch tensors. A custom DataLoader is employed to facilitate efficient model training by batching the data.

\textbf{Loss Function and Training Dynamics}

A salient characteristic of our modeling is creating a custom loss function. This function is rooted in a combination of Mean Squared Error (MSE) and a physics-based loss. Physics-based discrepancies are extracted from certain relationships inherent in the dataset, particularly focusing on the current density, Tafel slope, and exchange current density (known formula in Equation \ref{eq:uloss}). By combining the traditional MSE with the physics-based loss, the loss function imbibes the essence of both data-driven and domain-specific discrepancies. These were conceptualized to provide variations in computing the discrepancy from the physical equation, presenting potential avenues for further exploration and enhancement.

The PINN training regime is designed to optimize the network over specified epochs, with the custom loss ensuring adherence to both empirical data and underlying physical principles. An early stopping mechanism is built into the regime, preventing unnecessary computational effort in scenarios of stagnant or deteriorating validation loss.

A traditional ANN training function serves as a benchmark and basis for comparison. This function follows standard backpropagation practices, guided by MSE loss and supplemented by early stopping based on validation loss trends.

\textbf{Variation of Parameters}

Consistent with the manuscript's assertion, we extensively experiment with the value of alpha in the custom loss function. This parameter is pivotal, in determining the weight accorded to the MSE and the physics-based components of the loss function. By varying alpha values between 0.1, 0.5, 1, 2, and 5, we aim to discern the optimal balance that ensures predictions are accurate while staying true to domain knowledge.

Furthermore, our experiments incorporate partial knowledge integration. This entails the integration of various variables and operators from the governing formula, ensuring a comprehensive understanding of their collective influence on the model's predictions. The iterative approach, spanning 20 rounds for each configuration, serves to furnish robust statistical conclusions regarding performance metrics.

The final evaluation phase computes the \textbf{normalized root mean squared error (NRMSE)} for each model to measure their predictive performance against actual observations.

\section{Knowledge Discovery Configuration - Level 3}
\label{appendix:l3}
In the provided $\phi$-SO training configuration, symbolic operations include multiplication, addition, subtraction, division, and various mathematical functions such as inverse, squared, square root, exponential, logarithmic, sine, and cosine functions. The defined input variables being considered are current density ($i$), exchange current density ($i_0$), and Tafel slope ($A$) with their respective units specified in relation to the international system of units [$L$, $M$, $T$, $I$, $\theta$, $N$, $J$] as the constraint, which stands for length, mass, time, electric current, thermodynamic temperature, amount of matter, and luminous intensity, respectively.

The symbolic modeling is configured such that it takes the above arguments and aims to generate equations with a super-parent 'Activation Losses'. The training adopts a reward mechanism to ensure the physicality of the results, where the SquashedNRMSE function \cite{tenachi2023deep} assesses the rewards. The learning configuration prescribes the usage of a batch size of 200 and a maximum length of 35 for the symbolic equations. The training employs the Adam optimizer with a learning rate of 0.0025 for model parameter optimization.

During the training process, several priors are imposed. These include setting a range for the length of the generated equations, preventing unnecessary inverse operations, ensuring physical units, constraining the nesting of trigonometric functions, and limiting occurrences of certain terms. The Long Short-Term Memory cell \cite{hochreiter1997long} configuration has a hidden size of 128 and one layer. The performance monitoring metric during the process of the symbolic formula is measured by the Coefficient of determination.

After training, a Pareto front is constructed using the logged results, plotting the relationship between the complexity of the symbolic equation and its RMSE. The symbolic equations obtained from the Pareto front are then showcased, along with the determined values of their free constants.

\bibliographystyle{unsrt}  
\bibliography{references}

\end{document}